\documentclass[12pt]{article}
\usepackage{amssymb,amsmath,cite,bm}
\usepackage[dvips]{graphicx,color}
\usepackage{psfrag,subfigure}
\begin{document}

\title{\bf Seamless Capture and Stabilization of Spinning Satellites By Space Robots with Spinning Base}

\author{Farhad Aghili\thanks{email: farhad.aghili@concordia.ca}}

\date{}

\maketitle

\begin{abstract}
This paper introduces an innovative guidance and control method for simultaneously capturing and stabilizing a fast-spinning target satellite, such as a spin-stabilized satellite, using a spinning-base servicing satellite equipped with a robotic manipulator, joint locks, and reaction wheels (RWs). The method involves controlling the RWs of the servicing satellite to replicate the spinning motion of the target satellite, while locking the manipulator's joints to achieve spin-matching. This maneuver makes the target stationary with respect to the rotating frame of the servicing satellite located at its center-of-mass (CoM), simplifying the robot capture trajectory planning and eliminating post-capture trajectory planning entirely. In the next phase, the joints are unlocked, and a coordination controller drives the robotic manipulator to capture the target satellite while maintaining zero relative rotation between the servicing and target satellites. The spin stabilization phase begins after completing the capture phase, where the joints are locked to form a single tumbling rigid body consisting of the rigidly connected servicing and target satellites. An optimal controller applies negative control torques to the RWs to dampen out the tumbling motion of the interconnected satellites as quickly as possible, subject to the actuation torque limit of the RWs and the maximum torque exerted by the manipulator's end-effector.
\end{abstract}

\section{Introduction}
Over the past few decades, there has been a growing interest in on-orbit servicing (OOS) to encompass a broader range of tasks, such as maintenance and repair, rescue operations, refueling, inspection, and even orbital debris removal~\cite{Hirzinger-Landzettel-Brunner-Fisher-Preusche-2004,Yoshida-Dimitrov-Nakanishi-2006,Giordano-Ott-2019,Papadopoulos-Aghili-2021}. Many space agencies have identified autonomous robotic systems as the key technology for on-orbit servicing missions~\cite{Goddard-2010}. The common guidance and control strategy for autonomous space robots to capture spinning satellites can be divided into two phases: the pre-capturing phase and the post-capturing phase. During the pre-capturing phase, the robot must intercept a grapple fixture on the spinning target with zero relative velocity to minimize impact, while in the post-capturing phase, the robotic system must bring the tumbling satellite to rest. Although many research works have addressed exclusively the guidance and control problem in either the pre-capturing or post-capturing phase, seamlessly integrating robotic planning and control in both phases while satisfying multiple constraints remains a challenging task.

Robot motion-planning and guidance techniques for intercepting a spinning target in the literature can be classified into two categories: robot-motion planning and guidance for the pre-capturing phase and robot-motion planning and guidance for the post-capturing phase \cite{Aghili-2009d,Zhanga-Lianga-2017,Aghili-2023}. During the pre-capturing phase, the manipulator's arm must be guided, typically using vision data, so that its end-effector intercepts the satellite grapple at a rendezvous point along the trajectory of the target satellite in such a manner that the relative velocity between the end-effector and the target grapple point is zero at the time of capture to minimize impact. In the post-capturing phase, the space manipulator should gently exert torque on the grasped spinning satellite to stop its angular motion. Recent surveys of research work done in the area of robotic trajectory planning and capture in space can be found in  \cite{Papadopoulos-Aghili-2021}.
The existing body of research on path-planning and control of servicing space robots assumes that the base attitude of the servicing satellite must be stabilized using either coordination control or reaction null-space based control when the robotic manipulator is guided to intercept a target \cite{Giordano-Ott-2019,Aghili-2011k,Lampariello-Hirzinger-2013,Hirano-Kato-Tanishima-2017,Aghili-Parsa-2009b}. Trajectory planning and control schemes for intercepting a grapple point on a spinning satellite are described in \cite{Oki-Nakanishi-Yoshida-2008,Aghili-Parsa-2007b}. After capturing a noncooperative target, a coordinated stabilization control for a space robot is proposed in \cite{Zhanga-Lianga-2017} that allows for the reduction of angular momentum in two stages of Momentum Reduction and Momentum Redistribution. In \cite{Chu-Wu-2018,Aghili-Namvar-2008}, a self-learning soft-grasp control algorithm based on variable stiffness technology for target capturing by a free-floating space manipulator is described. Additionally, there are several studies on path-planning and control schemes for the post-capturing phase aimed at bringing the target satellite to rest \cite{Aghili-2008c,Dimitrov-Yoshida-2004,Aghili-2019e}. Recent trajectory planning and coordination control schemes for detumbling a non-cooperative target with unknown inertial parameters can be found in \cite{Huang-Lu-Wang-2019,Aghili-2019c,Liu-Liu-2021,Aghili-2020a,Gangapers-Liu-2021,Namvar-Aghili-2003}. Furthermore, a real-time obstacle avoidance strategy for a redundant space manipulator in null space is proposed in \cite{Li-Yang-2022}. Despite progress in the guidance and control of space robots, the seamless end-to-end trajectory planning and control for both pre- and post-capturing phases remain challenging.

This work proposes a novel approach for capturing and stabilizing a spinning satellite using a servicing space robot with a spinning base. As shown in Fig.~\ref{fig:3phases}, the entire operation consists of three phases: $i)$ matching spin maneuver, $ii)$ simultaneous grasping and spin synchronization, and $iii)$ detumbling by transferring the angular momentum to the reaction wheels. It is assumed that the spin axis of the target satellite remains unchanged, which is typical for spin-stabilized satellites, and that the center of mass (CoM) of the servicing satellite is initially placed along with the axis of rotation of the target. By matching the spin between the two satellites, the trajectory planning in the pre-capture phase is simplified, and any robot trajectory planning in the post-capture phase is unnecessary as the robot's joints are locked. Therefore, only one robot trajectory plan suffices for both pre-capture and post-capture phases. During the pre-capture phase, the attitude controller of the servicing satellite maintains spin synchronization, while during the post-capture phase, it stabilizes the rotation motion of the rigidly interconnected servicing and target satellites, subject to various constraints such as the actuation torque limit of the reaction wheels and the maximum torque that can be exerted by the manipulator's end-effector.

\section{Guidance and Control} \label{sec:Multi-axis}

\begin{figure*}
\psfrag{Target satellite}[c][c][.7]{Target satellite}
\psfrag{Servicing satellite}[c][c][.7]{Servicing satellite}
\psfrag{locked joints}[c][c][.7]{Locked joints}
\psfrag{Phase A}[c][c][.8]{Phase A}
\psfrag{Phase B}[c][c][.8]{Phase B}
\psfrag{Phase C}[c][c][.8]{Phase C}
\psfrag{wt}[l][l][.8]{$\bm\omega_s$}
\psfrag{wt}[l][l][.8]{$\bm\omega_s$}
\psfrag{wb}[l][l][.8]{$\bm\omega_{b}$}
\psfrag{tb}[l][l][.8]{${\bm\tau}_{r}$}
\psfrag{vrho}[l][l][.8]{$\bm\varrho$}
\psfrag{rho}[l][l][.8]{$\bm\rho$}
\psfrag{rb}[l][l][.8]{$\bm r_b$}
\psfrag{rf}[l][l][.8]{$\bm r_s$}
\psfrag{tm}[l][l][.8]{$\bm\tau_m$}
\psfrag{Et}[l][l][.7]{$\Sigma_S$}
\psfrag{Eb}[l][l][.7]{$\Sigma_B$}
\psfrag{Eg}[l][l][.7]{$\Sigma_C$}
\psfrag{r}[c][c][.8]{$\bm r$}
\psfrag{teti}[c][c][.8]{$\bm\theta_i$}
\psfrag{tetf}[c][c][.8]{$\bm\theta_f$}
\psfrag{rs}[c][c][.8]{$\bm r_s$}
\psfrag{c}[c][c][.8]{$\bm\varrho$}
\psfrag{CoM}[c][c][.7]{CoM}
\psfrag{rs}[c][c][.8]{$\bm r_s$}
\centering {\includegraphics[clip,width=16cm]{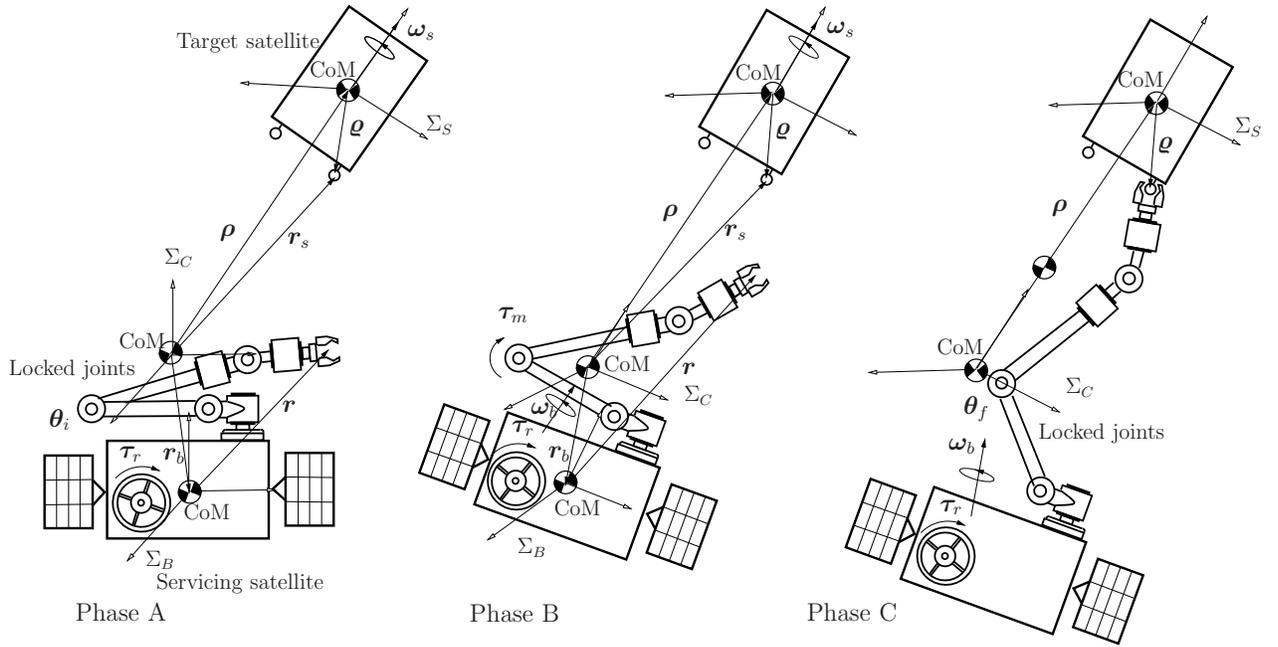}}
\caption{The concept of capture and stabilization of a spinning space target  by a spinning-base servicing robot.} \label{fig:3phases}
\end{figure*}

Fig.~\ref{fig:3phases} illustrates schematically the innovative concept of robotic capture and stabilization of a spinning satellite in three sequentially occurring phases. We assume the servicing satellite is equipped with an articulated manipulator and momentum wheels that will be used for grasping a target satellite and storing its angular momentum, respectively. The manipulator joints are equipped with friction based locking devices which can  hold rigidly the position of the joints upon command without requiring active torque control. We also assume that the servicing satellite is positioned such that its CoM is aligned with the rotation axis of the target and that the rotation axis of the target is not changed, i.e., spin-stabilized spacecraft. Three frames are defined as follows: The base frame, $\Sigma_B$, is attached to the base of the servicing satellite and positioned at its CoM. The body frame, $\Sigma_B$, is attached to the target satellite at its CoM and is aligned with its principal axes. Frame $\Sigma_C$ is placed at the CoM of entire servicing satellite, including both the robotics arm and the base, and its axes aligned with those of $\Sigma_B$. It should be noted that $\Sigma_C$ becomes a rotating frame if the base frame $\Sigma_B$ rotates.    

In phase A, the angular rate and the orientation of the servicing satellite's base are matched to those of the  target  through control action of the reaction wheels thereby eliminating the relative angular motion between the servicing and target satellites. All robot joints are locked during the matching spin maneuvers and hence no joint control is required for this phase. Subsequently, in phase B, the robot's joints are unlocked and a coordination controller drives the robot to capture  a fixture point on the target while maintaining zero relative angular velocity between the two spacecraft. In this phase  the robot joints are locked again in order to form a single rigid-body system comprising the interconnected servicing and target satellites, which continues to rotate due to conversation of angular momentum. Control torques are applied on the momentum wheels and in the base of the servicing satellite in opposite direction in order to transfer the angular momentum of the satellite body to the reaction wheels in order to  bring the rigidly interconnected satellites to rest as quickly as possible given torque limits of the wheels actuators torque and the end-effector.

The dynamics equations governing the motion of a free-flying space robot as a multibody system can be expressed in the following form ~\cite{Xu-Kanade-1993,Dimitrov-Yoshida-2004}: 
\begin{equation} \label{eq:general_robot_dynmamics}
\begin{bmatrix} \tilde{\bm M}_{b}  & \bm M_{bm}  & \bm M_{br}  \\ \bm M^T_{bm}  & \bm M_{m}& \bm 0 \\ \bm M_{br}^T & \bm 0 & \bm M_r \end{bmatrix}
 \begin{bmatrix}\dot{\bm\omega}_b  \\ \ddot{\bm\theta} \\\ddot{\bm\phi}  \end{bmatrix} +  \begin{bmatrix} \tilde{\bm c}_b \\ \bm c_m \\ \bm c_r  \end{bmatrix}
 = \begin{bmatrix} \bm 0 \\ \bm\tau_m \\ \bm\tau_{r} \end{bmatrix} + \begin{bmatrix} {\bm J}_b^T \\ \bm J_m^T \\ \bm 0 \end{bmatrix} \bm n_e
\end{equation}
where $\tilde{\bm M}_{b}$, $\bm M_r$, $\bm M_r$ represent the inertia matrices associated with the satellite base, manipulator, and reaction wheels, $\bm M_{bm}$ and $\bm M_{br}$ denote the cross-coupling  inertia matrices, ${\bm J}_b$ and $\bm J_m$ are the  submatrices of the manipulator's Jacobian with respect to its end-effector, $\bm n_e^T=[\bm f_e^T \; \bm\tau_e^T]$ is the  applied wrench to the manipulator end-effector,  vector $\bm\theta$ corresponds to manipulator's joint angles, $\bm\omega_{b}$ is the angular velocity of the base of the servicing satellite expressed in coordinate frame $\Sigma_B$,  $\bm\tau_m$ is the manipulator's joint torques, $\bm\tau_{r}$ is the actuator torques of the reaction wheels, and nonlinear vectors $\tilde{\bm c}_{b}$, $\bm c_m$, $\bm c_r$ collectively encapsulate the generalized Coriolis and centrifugal terms~\cite{Xu-Kanade-1993}. Note that that the zero element in the vector of generlaized force in \eqref{eq:general_robot_dynmamics} is because no force or torque is applied to the based on the servicing satellite. We can reduce the dimension of the matrix equation in \eqref{eq:general_robot_dynmamics} by substituting the wheel acceleration $\ddot{\bm\phi}$ into the upper equations. This reduction implicitly incorporates $\ddot{\bm\phi}$, yielding a more concise representation, i.e.,
\begin{equation} \label{eq:robot_dynmamics}
\underbrace{\begin{bmatrix} {\bm M}_{b}  & \bm M_{bm} \\ \bm M_{bm}^T & \bm M_m \end{bmatrix}}_{\bm M(\bm\theta)}
 \begin{bmatrix}  \dot{\bm\omega}_b \\ \ddot{\bm\theta} \end{bmatrix} +  \begin{bmatrix} {\bm c}_{b} \\ \bm c_m \end{bmatrix}
 = \begin{bmatrix} \bm B \bm\tau_r \\ \bm\tau_m \end{bmatrix} + \begin{bmatrix}\bm J_b^T \\ \bm J_m \end{bmatrix} \bm n_e 
\end{equation}
where  $\bm M(\bm\theta)$ is the generalized inertia matrix of the servicing satellite and
\begin{align*} \notag
\bm B & = -\bm M_{br} \bm M_r^{-1}\\
\bm M_{b} & = \tilde{\bm M}_b + \bm B \bm M_{br} \\
{\bm c}_{b} &=\tilde{\bm c}_b + \bm B \bm c_r
\end{align*}

\subsection{Phase A: Matching Spin Maneuver (MSM)} \label{sec:MSM}
This section describes a control law for attitude maneuvers of the servicing robot so that  the  angular rate and the orientation of the robot's base instantaneously match those of the target satellite. During the matching spin maneuver, all joints of  the robotic manipulator are locked and hence the servicing satellite can be considered as a single rigid-body system. Assuming $\bm\omega_b$ and $\bm\omega_s$ represent the angular velocities of the servicing  and target satellites expressed in their body-fixed coordinate frames $\Sigma_B$ and $\Sigma_S$, respectively, we can write the equation of the  relative angular velocity in the rotating frame $\Sigma_B$ by
\begin{equation} \label{eq:omega_rel}
\bm \omega_{\rm rel} = \bm\omega_b - \bm A(\bm q) \bm\omega_s.
\end{equation}
Here, $\bm A$ is the rotation matrix corresponding to quaternion $\bm q$ representing the orientation of frame $\Sigma_S$ with respect to frame $\Sigma_B$. The rotation matrix as a function of the quaternion is given by
\begin{equation} \label{eq:R}
\bm A(\bm q) =\bm I + 2 q_o [\bm q_v \times] + 2 [\bm q_v \times]^2,
\end{equation}
where  $\bm q_v$ and $q_o$ are the vector and scalar parts of the
quaternion, i.e., $\bm q=[\bm q_v^T \; q_o]^T$, $[\cdot \times]$ denotes the matrix form of the cross-product,
and $\bm I$ denotes  identity matrix with adequate dimension.  We assume  the target is being axisymmetric rigid body, i.e., $I_{xx}=I_{yy}$, and thus the its inertia matrix takes the form $\bm I_c = I_{xx} \mbox{diag} \{1 , 1, 1- \lambda \}$ where $\lambda =1 - I_{zz}/I_{xx}$. In this case,   
one can write the equations of the target by
\begin{equation} \label{eq:Mtwt}
\dot{\bm\omega}_s =  \bm\phi(\bm\omega_s)    \quad \mbox{where} \quad \bm\phi(\bm\omega_s) = \lambda  \begin{bmatrix} \omega_{y_s} \omega_{z_s} \\  -\omega_{x_s} \omega_{z_s}  \\ 0 \end{bmatrix},
\end{equation}
where $\omega_{x_s}$ and  $\omega_{y_s}$ are the components of the  angular velocity causing  change in the orientation of the rotational axis of the rotating body and $\omega_{z_s}$ is the axial angular velocity. We further assume zero transverse angular velocity and thus constant angular velocity vector $\bm\omega_s = \omega_{z_s} \hat{\bm k}$ with $\hat{\bm k}=[0\; 0 \; 1]^T$ that is the case of a spin-stabilized satellite. 
 
We assume that the servicing satellite is placed with zero relative velocity with respect to the target satellite. As no external forces act upon the system of the servicing satellite and the target, the displacement vector connecting their centers of mass remains constant and is unaffected by the movement of the robotic arm due to the conservation law of linear momentum in space. Let us define displacement vectors $\bm{\rho}$ and $\bm\varrho$, which represent the locations of the servicing satellite CoM and the grasping point, respectively, both expressed in the target body frame $\Sigma_S$, see Fig.~\ref{fig:3phases}. Since the relative liner velocity of two satellites is zero, the velocity of the   
grasping point on the target expressed in the rotating frame $\Sigma_C$ can be written as 
\begin{equation}
\bm v_s= \bm\omega_s \times \bm\rho
\end{equation}
provided  that the attitudes of both satellites are instantaneously aligned, i.e., $\bm A \equiv \bm I$ and $\bm\omega_b = \bm\omega_s$. This implies that aligning the displacement vector $\bm\rho$ with the rotation axis of the target, i.e., $\bm\rho \parallel \bm\omega_s$, is necessary to zero the relative velocity. Therefore, we assume that the servicing satellite is initially positioned with respect to the target in a manner that satisfies the latter condition. With the instantaneous alignment conditions established, we can now proceed with the control design. Suppose $\bm\theta_i$ represents the joint angles during Phase A, when all joints are locked. Since  $\ddot{\bm\theta}_i = \dot{\bm\theta}_i \equiv \bm 0$, we have $\bm M_{b}=\bm M_b(\bm\theta_i) \mbox{const}$ as a constant matrix. Thus, the dynamics equation of the servicing satellite can be simplified to:
\begin{equation} \label{eq:Mbwb}
\bm M_{b} \dot{\bm\omega}_b + \bm c_b = \bm B \bm\tau_{r},
\end{equation}
where $\bm h_{r}$ is the angular momentum delivered by the reaction wheels. 
Substituting the angular velocity  $\dot{\bm\omega}_b$ from  \eqref{eq:Mbwb} into the time-derivative of \eqref{eq:omega_rel}, we arrive at the following expression of the relative acceleration
\begin{equation} \label{eq:dot_omega_rel}
\dot{\bm \omega}_{\rm rel} = - \bm\omega_{\rm rel} \times \bm A(\bm q)\bm\omega_s  + \bm M_{b}^{-1} \big( - \bm c_r  + \bm B \bm\tau_{r} \big).
\end{equation}
Suppose unit quaternion ${\bm q}^*=[0 \; 0 \; 0 \; 1]^T$ represent the desired orientation of the two satellites, which is   corresponding to $\bm A(\bm q^*) = \bm I$. 
Now consider the following feedback for the servicing attitude control
\begin{equation} \label{eq:contr_law1}
\bm\tau_{r}  = -\bm B^{-1} \Big( \bm c_b  +\bm M_{b} \big( k_p  \bm q_v   + k_d \bm\omega_{\rm rel} \big) \Big)
\end{equation}
where  scalars $k_p>0$ and $k_d>0$ are the feedback gains, which are with dimensions ${\rm s}^{-2}$ and ${\rm s}^{-1}$.  In the following analysis, we will show that control law \eqref{eq:contr_law1} can modify the angular motion dynamics of the base so that its angular velocity and orientation  asymptomatically track those of the target satellite.
Substituting the control law \eqref{eq:contr_law1} into \eqref{eq:dot_omega_rel} will result in the following
differential equations
\begin{equation} \label{eq:dot_omeg_dyn}
\dot{\bm \omega}_{\rm rel} = - \bm\omega_{\rm rel} \times \bm A(\bm q) \bm\omega_s - k_p  {\bm q}_v - k_d \bm\omega_{\rm rel},
\end{equation}
Define the following positive-definite Lyapunov function:
\begin{equation}
V =   \frac{1}{2} k_p \| \bm q - {\bm q}^* \|^2  + \frac{1}{2} \| \bm\omega_{\rm rel} \|^2.
\end{equation}
Also, consider the following identity which relates the time-derivative of the quaternion to the angular velocity
\begin{equation}\label{eq:dot_q}
\dot{\bm q} = \frac{1}{2} \bm\Omega(\bm\omega_{\rm rel}) \bm q, \quad \mbox{where} \quad   \bm\Omega(\bm\omega)\triangleq \begin{bmatrix} -[\bm\omega  \times]   & \bm\omega \\
- \bm\omega^T & 0
\end{bmatrix}.
\end{equation}
Then, using   \eqref{eq:dot_q} in  the time-derivative of $V$ along trajectories \eqref{eq:dot_omeg_dyn}, we will arrive at
\begin{align} \notag
\dot V = & k_p (\bm q - {\bm q}^* )^T \dot{\bm q} - \bm\omega_{\rm rel}^T  \dot{\bm\omega}_{\rm rel}\\ \notag
= & \frac{1}{2}k_p \big( (\bm q - {\bm q}^*)^T \bm\Omega(\bm\omega_{\rm rel}) \bm q - \bm\omega_{\rm rel}^T  \Big( \bm\omega_{\rm rel} \times \bm A(\bm q) \bm\omega_s \Big) \\ \label{eq:dotV1}
&- k_d \| \bm\omega_{\rm rel} \|^2  - k_p \bm\omega_{\rm rel}^T {\bm q}_v
\end{align}
By inspection, one can verify the following useful identities
\begin{subequations} \label{eq:identities}
\begin{equation}
(\bm q - \bm q^*)^T \bm\Omega(\bm\omega_{\rm rel}) \bm q   =  \bm\omega_{\rm rel}^T {\bm q}_v
\end{equation}
\begin{equation}
\bm\omega_{\rm rel}^T  \Big( \bm\omega_{\rm rel} \times \bm A(\bm q) \bm\omega_s \Big)  = \bm 0
\end{equation}
\end{subequations}
Finally, using \eqref{eq:identities} in \eqref{eq:dotV1} yields
\begin{equation}
\dot{V} = - k_d \| \bm\omega_{\rm rel} \|^2 \leq 0.
\end{equation}
Therefore, according to the LaSalle's Global Invariant Set Theorem
\cite{Khalil-1992-p115}, the equilibrium point reaches
where $\dot V=0$, or $\bm\omega_{\rm rel} \equiv \bm 0$. Then, by virtue of \eqref{eq:dot_omeg_dyn}  we can infer  identity ${\bm q}_v =0$, which implies  global asymptotic convergence of the orientation error, i.e.,  $\bm\omega_{\rm rel} \rightarrow 0$ and $\bm A(\bm q) \rightarrow \bm I$ as $t\rightarrow \infty$.

\subsection{Phase B: Simultaneous  Grasping and Spin Synchronization} \label{sec:grasping}
Once the matching spin maneuvers are complete, the manipulator's joints are unlocked to initiate the grasping phase. Our goal now is to develop a coordination control scheme that will enable the robotic manipulator to move from its parked position $\bm\theta_i$
to the grasping fixture on the target satellite while maintaining attitude synchronization. This will ensure that the target remains stationary with respect to the servicing satellite frame $\Sigma_C$. To accomplish this, we use the following notation: $m_b$ and $\bm r_b$ denote the mass of the servicing satellite base and its location of CoM expressed in the frame $\Sigma_C$, while $m_{i}$ and $\bm r_{c_i}$ denote the mass of the $i$-th link and its location of CoM in the body frame $\Sigma_C$. Using the definition of the CoM of the overall multibody system, we obtain the following equations:
\begin{equation}
\bm r_b(\bm\theta) = - \frac{1}{m_b} \sum_i m_{i} \bm r_{c_i}(\bm\theta),
\end{equation}
\begin{equation} \label{eq:Jb_dot_theta}
\dot{\bm r}_b = \bm J \dot{\bm\theta},
\end{equation}
where $\bm J = \partial \bm r_b(\bm\theta) / \partial \bm \theta$ is the cosponsoring Jacobian. 
Suppose vectors $\bm r(\bm\theta)$ and $\bm\eta(\bm\theta)$ denotes the position and orientation of the robot end-effector expressed in the base  frame $\Sigma_B$. Suppose $\bm\theta_f = \bm\theta(t_f) $ denotes the joint angles when the robotic arm reaches the grasping point at terminal time $t_f$, and thus $\dot{\bm\theta}_f = \dot{\bm\theta}(t_f) =  \bm 0$. Form the system kinematics illustrated in Fig.~\ref{fig:3phases}, if the frames of two satellites are aligned then the following vector equation holds 
\begin{equation} \label{eq:r_f}
\bm r(\bm\theta_f) = \bm\rho  + \bm\varrho + \bm r_b(\bm\theta_f),  
\end{equation}  
Note that frame $\Sigma_C$ is located at the center of mass (CoM) of the entire servicing satellite, comprising the robotic arm and the base. As per the conservation law of linear momentum in space, the position of the CoM, also referred to as the virtual ground, remains unaffected by the arm's movement, resulting in a constant position vector $\bm\rho$.   
As will be later described, a  coordination controller is responsible for  the angular   alignment between two satellites in Phase B, i.e., $\bm A \equiv \bm I$ and hence both  $\bm r_s=\bm\rho +   \bm\varrho$. In other words, the pose of the grapple fixture $\bm\chi_s$ expressed in  $\Sigma_G$ remains constant  by virtue of the fact that the movement of the manipulator does not change the CoM position of the servicer satellite while the orientation of the servicer's base is always aligned with that of the target satellite through a coordination control. That is instantaneous angular alignment of the frame $\Sigma_S$ with parallel frames $\Sigma_B$ and $\Sigma_C$. Under this circumstance,  one can derive the final joint angles $\bm\theta_f$ algebraically by solving the displacement equation \eqref{eq:r_f} along with the desired orientation of the end-effector at the grasping point, as specified by $\bm\eta(\bm\theta_f) = \bm\eta_f$. The kinematic equations can be solved for $\bm\theta_f$ using analytical or numerical methods \cite{Goldenberg-1985}. Subsequently, the joint trajectories $\bm\theta^*(t), \dot{\bm\theta}^*(t), \ddot{\bm\theta}^*(t) $  planned for  time interval $[0, \; t_f ]$  pertaining to Phase B can be obtained from
\begin{equation} \label{eq:theta_star}
\begin{bmatrix} \bm\theta^*(t) \\ \dot{\bm\theta}^*(t) \\ \ddot{\bm\theta}^* (t)\end{bmatrix} =
\begin{bmatrix}  \bm\theta_i +  ( 3 \hat{t}^2  - 2 \hat{t}^3  ) \Delta \bm\theta \\ (6 \hat{t} - 6 \hat{t}^2 )  \Delta \bm\theta  /{t_f} \\ (6  - 12 \hat{t} ) \Delta \bm\theta /t^2_f \end{bmatrix},
\end{equation}
where  $\Delta \bm\theta = \bm\theta_f - \bm\theta_i$ and $\hat{t} = t/t_f$ is the normalized time. Suppose the maximum allowable joint rate and acceleration are denoted by $\dot \theta_{\rm max}$ and $\ddot \theta_{\rm max}$, respectively. Since the maximum joint rate  and maximum joint acceleration occur at the normalized time $\hat{t} = 1/2$ and  $\hat{t} = 0$, the minimum terminal time $t_f$ satisfying the maximum joint rate and acceleration constraints can be determined from
\begin{equation} \label{eq:tf}
t_f = \max \left\{ \frac{3}{2 \dot\theta_{\rm max}} \| \Delta \bm\theta \|_{\infty} , \quad \Big( \frac{6}{ \ddot\theta_{\rm max}} \| \Delta \bm\theta \|_{\infty} \Big)^{1/2} \right\} 
\end{equation}
where $\| \Delta \bm\theta \|_{\infty} = \max_j | \Delta \theta_j |$ denotes the maximum norm of a vector. 
Substituting the terminal time from \eqref{eq:tf} into \eqref{eq:theta_star} yields the joint trajectories from the parked position $\bm\theta_i$ to the final position $\bm\theta_f$ corresponding to the grasping point. This ensures that the maximum joint rates and acceleration remain at their maximum values, and the joint rate at the terminal time is zero, i.e., $\dot{\bm\theta}_f = \bm 0$. The latter equality, together with the virtue of \eqref{eq:Jb_dot_theta}, implies that $\dot{\bm r}_b(t_f) = \bm 0$, indicating zero relative velocity between the frames $\Sigma_B$ and $\Sigma_C$ at the instance of capture time $t_f$. In other words, the condition for smooth capture is satisfied since the robotic end-effector reaches the grasping point at terminal time $t_f$ with zero relative velocity. To move the robotic arm according to the joint trajectories dictated by \eqref{eq:theta_star} and simultaneously spin the base to be instantaneously aligned with the attitude of the target satellite,  For this purpose, we consider the following coordination control law.
\begin{align} \notag
\begin{bmatrix}  \bm\tau_{r} \\ \bm\tau_m\end{bmatrix} = & \begin{bmatrix} \bm B^{-1}\bm c_{r} \\ \bm c_m \end{bmatrix} + \begin{bmatrix} \bm M_r \bm M_{br}^{-1} \bm M _b & \bm M_r \bm M_{br}^{-1} \bm M_{bm} \\ \bm M^T_{bm}  & \bm M_m \end{bmatrix} \\ \label{eq:coord_control} & \qquad \times \begin{bmatrix}   \bm\omega_{\rm rel} \times \bm A \bm\omega_s +     k_{\omega} \bm\omega_{\rm rel}   + k_q {\bm q}_v  \\ \ddot{\bm\theta}^* + k_d (\dot{\bm\theta}^* - \dot{\bm\theta}) +  k_p (\bm\theta^* - \bm\theta) \end{bmatrix}
\end{align}
where   $\bm\theta^*, \dot{\bm\theta}^*, \ddot{\bm\theta}^*$ are obtained from \eqref{eq:theta_star}, and $k_p>0$, $k_d>0$, $k_{\omega}>0$, $k_q >0$ are the feedback gains. Substituting the expression of the control law from \eqref{eq:coord_control} into the system dynamics \eqref{eq:robot_dynmamics} yields a set of two uncoupled differential equations
\begin{equation}
\ddot{\tilde{\bm\theta}} + k_d \dot{\tilde{\bm\theta}} + k_p {\tilde{\bm\theta}} = \bm 0
\end{equation}
and
\begin{equation}
\dot{\bm\omega}_{\rm rel}  + k_{\omega} {\bm\omega}_{\rm rel} + k_q {\bm q}_v
\end{equation}
Stability of the above systems can be  proved using the standard argument, i.e., $\bm\theta \rightarrow \bm\theta^*$, $\bm A(\bm q) \rightarrow  \bm I$, and $\bm\omega_{\rm rel} \rightarrow  \bm 0$.

\subsection{Phase C: De-spinning Via Angular Momentum Transfer to Reaction Wheels} \label{sec:detumbling}
Upon completion of the grasping phase, the joints  are locked again so that target can be rigidly connected to the servicing satellite constituting a single tumbling rigid-body system. The control objective is  to damp out the tumbling motion of the connected servicing and target satellites  as quickly as possible given while the Euclidean
norms of the reaction wheel torque and the end-effector torque are restricted to be below their maximum  prescribed
values $\tau_{r_{\rm max}}$ and $\tau_{e_{\rm max}}$, i.e.,
\begin{equation}\label{eq:ineqaulity}
\| \bm\tau_{r} \| \leq \tau_{r_{\rm max}} \quad \mbox{and} \quad
\| \bm\tau_e \| \leq \tau_{e_{\rm max}} 
\end{equation}
This is achieved by applying control torque to the momentum wheels, which also exert equal torque in the opposite direction on the satellite body, thereby the angular momentum of the satellite body is transferred to the reaction wheels for stopping the tumbling motion.
The states of system during de-spinning phase are given by: $\dot{\bm\theta} \equiv \bm 0 \;$, $\bm\omega_{\rm rel} \equiv \bm 0 \;$ , $\dot{\bm\omega}_b \neq \bm 0$, and $\bm A \equiv  \bm I$.
Knowing that $\dot{\bm\omega}_b =  \dot{\bm\omega}_s$, one can write the equation of target satellite as
\begin{equation} \label{eq:ddot_omega_se}
\bm I_c \dot{\bm\omega}_b  + \bm I_c \bm\phi(\bm\omega_b) = -\bm J_s^T \bm n_e,
\end{equation}
where  $\bm n_e$ is the wrench  applied to the robot end-effector by the target spacecraft and $\bm J_s^T=[[\bm\varrho \times] \;\;\; \bm I]$. To formulate the equations governing the interconnected dynamics of both servicing and target satellites during the post-capturing phase, we need to eliminate the variable $\bm{n_e}$ from equations \eqref{eq:robot_dynmamics} and \eqref{eq:ddot_omega_se}  considering their physical connection. Then, once derive the equation for the combined system of the rigidly connected  satellites as a single rigid-body
\begin{equation} \label{eq:Mbtddot_omega}
\bm M_t \dot{\bm\omega}_b + \bm c_t = \bm B \bm\tau_{r}
\end{equation}
where $\bm M_t$ represents the equivalent inertia of rigidly connected servicing and target satellites together, i.e.,
\begin{equation} \notag
\bm M_t=\bm M_b + \bm J_b^T \bm J_s^+ \bm I_c 
\end{equation}
$\bm J_s^+ =(\bm J_s \bm J_s^T)^{-1} \bm J_s$ and  $\bm c_t = \bm c_b + \bm J_b^T \bm J_s^+ \bm I_c \bm\phi(\bm\omega_b)$.
On the other hand, equation \eqref{eq:ddot_omega_se} can be redefined exclusively in relation to the torque input $\bm\tau_e$. This is due to the fact that the linear acceleration of the CoM of the target satellite, following rigidization, is contingent upon rotational motion, expressed as $\bm a_o=\dot{\bm\omega}_b \times \bm\rho_s + \bm\omega_b \times (\bm\omega_b \times \bm\rho_s)$ where displacement vector $\bm\rho_s$ represents the location of target CoM with respect to the CoM of the combined servicing and target satellites. Consequently, the force exerted by the robot end-effector by the target satellite is
\begin{equation} \label{eq:fe}
\bm f_e = -m_s \big( \dot{\bm\omega}_b \times \bm\rho_s + \bm\omega_b \times(\bm\omega_b \times \bm\rho_s)   \big)
\end{equation}  
where $m_s$ denotes the mass of the target satellite. Using \eqref{eq:fe} in \eqref{eq:ddot_omega_se}, the latter equation can be reformulated as follows: 
\begin{equation}\label{eq:ddot_omegab} 
\bm M_s \dot{\bm\omega}_b + \bm c_s = -\bm\tau_e, 
\end{equation} 
where $\bm M_s = \bm I_c + m_s [\bm\varrho \times][\bm\rho_s \times]$ and  $\bm c_s =\bm I_c \bm\phi (\bm\omega_b) - m_s \bm\varrho\times \big(\bm\omega_b \times(\bm\omega_b \times \bm\rho_s) \big)$.

Now substituting the angular acceleration from \eqref{eq:ddot_omegab} into \eqref{eq:Mbtddot_omega}, we can establish the relationship between $\bm\tau_e$ and $\bm\tau_{r}$ as follows
\begin{subequations}  \label{eq:tbte}
\begin{equation}
\bm G \bm\tau_{r} + \bm c_g= \bm\tau_e
\end{equation}
where
\begin{align}
\bm G & =   -\bm M_s \bm M_t^{-1}  \bm B  \\
\bm c_g  &= \bm M_s \bm M_t^{-1} \bm c_t - \bm c_s  
\end{align}
\end{subequations}
The affine equation \eqref{eq:tbte} shows how the attitude torque $\bm\tau_{r}$ is transmitted to the end-effector torque $\bm\tau_e$ during the de-spinning maneuver. Equations \eqref{eq:Mbtddot_omega}  and \eqref{eq:tbte} completely characterize the system in the detumbling phase and hence they will be utilized for control purposes in the followings.

The control problem being considered here is how to drive the rigidly connected satellites from the  initial angular velocity
$\bm\omega_b(0)$ to rest in {\em minimum time} while the Euclidean norms of the reaction wheel torque and the end-effector torque are restricted to be below their maximum  prescribed values $\tau_{r_{\rm max}}$ and $\tau_{e_{\rm max}}$, respectively. A Lyapunov technique
will be presented in this section.
Consider the following nonlinear controller for the detumbling phase
\begin{equation} \label{eq:cntrlaw}
\bm\tau_{r} = \bm B^{-1} \Big( \bm c_t - \frac{\bm M_t \bm\omega_b}{\| \bm M_t \bm\omega_b \|} \sigma \Big),
\end{equation}
where  $\sigma>0$ is a positive scalar to be determined later. Substitution of the expression of the control law from  \eqref{eq:cntrlaw} into  \eqref{eq:Mbtddot_omega} leads to  the dynamics of the closed-loop system
\begin{equation} \label{eq:closed-loop}
\dot{\bm\omega}_b =  - \frac{\bm\omega_b}{\| \bm M_t \bm\omega_b \|} \sigma
\end{equation}
Now, consider the magnitude of the angular momentum as Lyapunov function
\begin{equation} \label{eq:V}
V \triangleq \| \bm M_t \bm\omega_b \|.
\end{equation}
Then, the time-derivative of the function along the solution of \eqref{eq:closed-loop}  satisfies
\begin{align*}
\dot V &= \frac{\bm\omega_b^T \bm M_t^2 \dot{\bm\omega}_b}{\| \bm M_t \bm\omega_b \|} =  -   \frac{\bm\omega_b^T \bm M_t^2 \bm\omega_b }{\| \bm M_t \bm\omega_b \|^2} \sigma\\
&= - \sigma <0
\end{align*}
This means that the  controller reduces the magnitude of
the angular momentum at the constant rate of $\sigma$. Therefore, $\sigma$ should be selected as large as possible subject to inequality constraints  \eqref{eq:ineqaulity}. Upon substituting the torque expression from the control law \eqref{eq:cntrlaw} into the aforementioned inequalities, one can formally articulate the problem of determining the optimal scalar $\sigma$ that satisfies the constraints \eqref{eq:ineqaulity} in the following encapsulated optimization programming:
\begin{subequations} \label{eq:maxsigma}
\begin{align}
\max\quad & \sigma \\  \label{eq:equl_const}
\mbox{subject to:} \quad & \bm B \bm\tau_r + \frac{\bm M_t \bm\omega_b}{\|\bm M_t \bm\omega_b \|} \sigma - \bm c_t =0 \\ \label{eq:inequl_const1}
& \bm\tau_r^T \bm\tau_r - \tau^2_{r_{\rm max}} \leq0  \\ \label{eq:inequl_const2}
& \bm\tau_r^T \bm G^T  \bm G \bm\tau_r
 + 2 \bm c_g^T \bm G \bm\tau_r + \| \bm c_g \|^2 - \tau^2_{e_{\rm max}} \leq 0 
\end{align}
\end{subequations}
which possesses a solution in closed form, as detailed in the Appendix. Finally, substituting the optimal value of the control parameter $\sigma$ from \eqref{eq:sigma} into \eqref{eq:cntrlaw} completes the control law. Thus, if the control law \eqref{eq:cntrlaw} is applied to the reaction wheels of the servicing spacecraft, then the angular momentum of the rotating bodies consisting of the servicing and target satellites decays to zero as quickly as possible while the magnitude of the toque vectors exerted on the wheels and robot end-effector remain below their maximum allowable limits.

\section{Simulation Results} \label{sec:simulation}

This section presents dynamic simulation results for end-to-end demonstration of the proposed robotic concept of operation for simultaneous capture and stabilization of a spinning satellite. These results are pertaining to the simulation of matching spin maneuver phase, simultaneous grasping and synchronization phase, and detumbling via angular momentum transfer phase according to pictorial sequence in Fig.~\ref{fig:3phases}. The overarching goal is to demonstrate that the end-to-end robotic operation  achieves the prescribed functional requirements defined in Section~\ref{sec:Multi-axis}. The 4th order Runge Kutta integration method with step time 0.001~s was used to perform the simulation in Matlab. The inertial parameters of the servicing and target satellites are listed in Tables \ref{tab:inertia_servicing} and \ref{tab:inertia_sarget}, respectively. The controller bandwidth is set to $\varpi=1.8$~rad/s.

\begin{figure}
\psfrag{wrel}[c][c][.8]{$\bm\omega_{\rm rel}$~(rad/s)}
\psfrag{q}[c][c][.8]{$\bm q$ (quaternion)}
\centering{\includegraphics[clip,width=8cm]{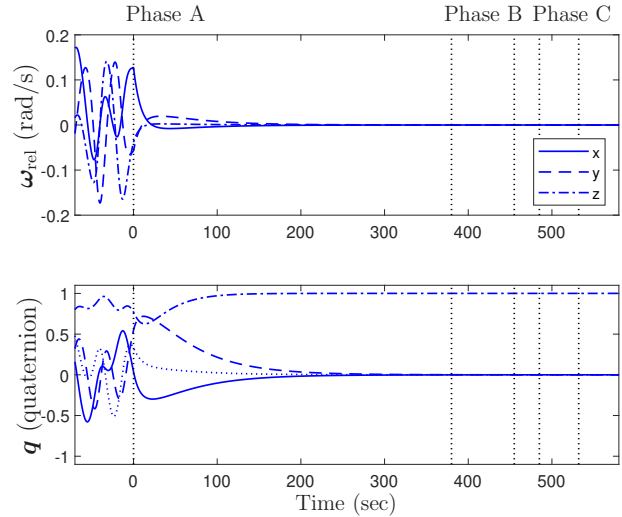}} \caption{The relative angular velocity and attitude of the target satellite w.r.t. the servicing satellite base.}\label{fig:omega_rel}
\end{figure}

\begin{figure}
\psfrag{r}[c][c][.8]{$\bm r_{\rm rel}$~(m)}
\psfrag{eta}[c][c][.8]{$\bm\eta_{\rm rel}$~(quaternion)}
\centering{\includegraphics[clip,width=8cm]{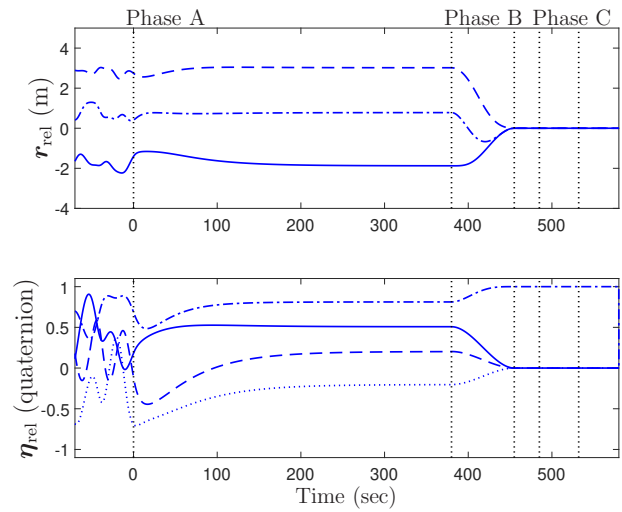}} \caption{The time-histories of the position and orientation of the target satellite w.r.t the robot end-effector.}\label{fig:pose_sarget_ee}
\end{figure}

\begin{figure}
\psfrag{v}[c][c][.8]{$\tilde{\bm v}_s$~(m/r)}
\psfrag{omg}[c][c][.8]{$\tilde{\bm\omega}_s$~(rad/s)}
\centering{\includegraphics[clip,width=8cm]{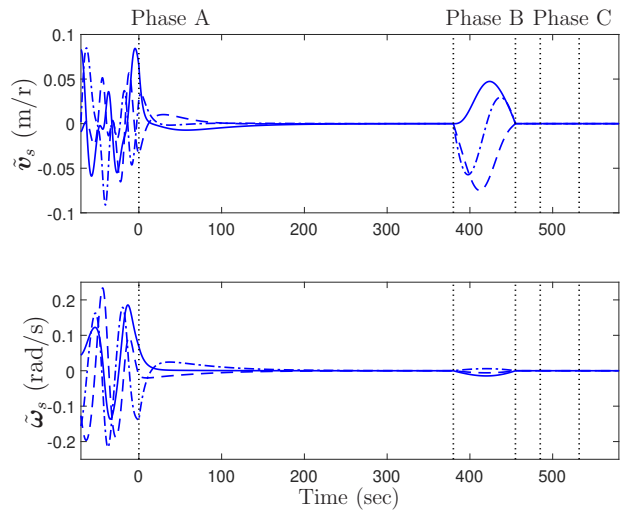}} \caption{The time-histories of the linear and angular velocities of the target satellite w.r.t the robot end-effector.}\label{fig:velocity_sarget_ee}
\end{figure}

\begin{figure}
\psfrag{taub}[c][c][.8]{$\bm\tau_{r}$~(Nm)}
\psfrag{taue}[c][c][.8]{$\bm\tau_e$~(Nm)}
\centering{\includegraphics[clip,width=8cm]{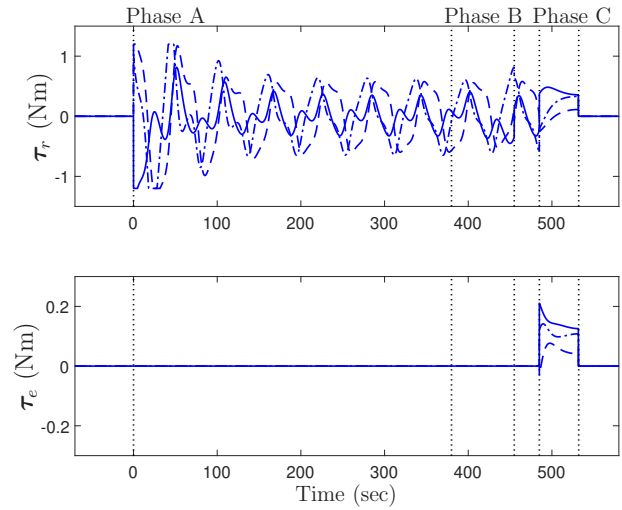}} \caption{The time-histories of the reaction wheels torque and end-effector torque.}\label{fig:torque_base}
\end{figure}

\begin{figure}
\psfrag{momentum}[c][c][.8]{Momentum~(Nm.s)}
\centering{\includegraphics[clip,width=8cm]{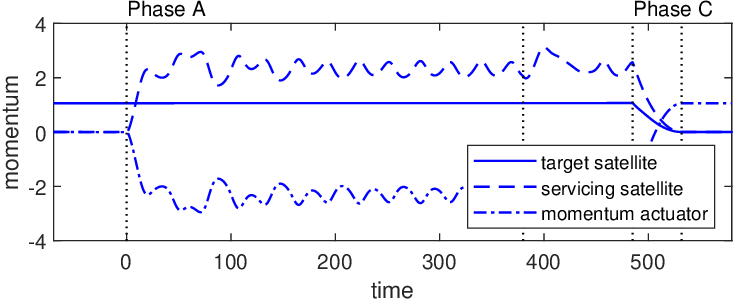}} \caption{The time-histories of the angular momentum magnitude of the target satellite versus that of the servicing satellite and its the momentum actuator.}\label{fig:momentum}
\end{figure}

The timing and sequence of events during the execution of the guidance and control of the servicing robot are listed in Table~\ref{tab:events}.  The time-histories of the relative angular velocity and attitude of the target satellite with respect to servicing satellite base are illustrated in  Fig.~\ref{fig:omega_rel}. It is apparent from the graphs that the spin synchronization control has succeeded  to annihilate the relative angular velocity and to align the orientation of two spacecraft at time $t_1=380$~s and thereafter. Trajectories of the target pose with respect to the manipulator's end-effector and the corresponding relative linear and angular velocities are plotted in  Figs.~\ref{fig:pose_sarget_ee} and \ref{fig:velocity_sarget_ee}, respectively. The plots demonstrate that the capture phase starts at time $t_1=380$~s and terminates successfully at $t_2=455$~s when the relative pose and velocities become zero. Upon completion of the capture phase, the joints are locked again and thus the entire system consisting of the servicing  and the target satellites virtually becomes a single rotating rigid-body. Next, the spin stabilization phase proceeds at time $t_3=485$~s by applying torque on the base of the servicing satellite and subsequently this phase terminates at time $t_1=538$~s, when the angular velocity gradually decays to zero as shown in  Fig.~\ref{fig:omega_base}. Trajectories of the torque applied to the servicing base and the robot end-effector are illustrated in  Fig.~\ref{fig:torque_base} for the entire operation including Phase A, B, and C. The control torques applied to the RW during Phase A and B  have been responsible to generate rotational motion of the base of the servicing satellite corresponding to that of the target satellite and then maintain the matching the spinning synchronization through the operation just prior to starting Phase C. The RW torques are applied from  $t_3=485$~s to $t_3=532$~s during Phase C in order to damp out the angular velocity of the rigidly connected servicing and target satellites. To this end, Fig.~\ref{fig:momentum} illustrates the magnitude of the angular momentum of the target  against that of the servicing satellite and momentum actuator. It is apparent from the graphs of the latter figure that  the initial momentum of the target satellite, which is $1.06~\mbox{Nm.s}$, has been precisely equal to the momentum stored in the momentum actuator at the end of Phase C. This is due to conversation of angular momentum in space. In other words, the coordinated robotic and attitude control maneuvers through Phases A, B, and C have succeeded to transfer the angular momentum of the target satellite to the momentum wheels of the servicing satellite.

\begin{table}
\caption{Timing and sequence of events.}
\begin{center}
\begin{tabular}{cccccc}
\hline \hline
Event   & synch.   & capture & capture & detumbling  & detumbling  \\
 & starts & starts & completes & starts & completes \\
 & $t_0$ & $t_1$ & $t_2$ & $t_3$  & $t_4$ \\
\hline
Time  & 0~s & 380~s & 455~s & 485~s & 532~s \\
\hline \hline
\end{tabular}
\end{center}
\label{tab:events}
\end{table}

\begin{table}
\caption{Inertial parameters of the servicing satellite.}
\begin{center}
\begin{tabular}{ccccc}
\hline
Body & Length  & CoM & Mass   & Inertia \\
     &  (m)  & (m) & (kg)    & (kg-m$^2$) \\
 \hline
Link 2,3   &  2   & 1 & 5  & $ \mbox{diag}[0.01, \; 2, \; 2]$ \\
Link 1,4,5,6   & 0.2 & 0.1  & 1  & $ \mbox{diag}[0.01, \; 0.02, \; 0.02]$ \\
Base & -         & - & 150 & $ \begin{bmatrix} 120 & 30  & -40 \\ 30 & 70 & 20 \\ -40 & 20 & 100  \end{bmatrix}$  \\
\hline
\end{tabular}
\end{center}
\label{tab:inertia_servicing}
\end{table}

\begin{table}
\caption{Inertial parameters of the target satellite.}
\begin{center}
\begin{tabular}{cc}
\hline
parameter & value\\
\hline
Mass (kg) & 200\\
CoM (m) &   $\bm\varrho=[-0.2 \;\; 0.1 \;\; -0.5 ]^T$ \\
Inertia (kg-m$^2$) &
$ \mbox{diag}[70, \; 70, \; 40]$ \\  \vspace{-3mm}\\
\hline
\end{tabular}
\end{center}
\label{tab:inertia_sarget}
\end{table}

\psfrag{time}[c][c][.7]{Time (sec)}
\psfrag{Phase A}[c][c][.7]{Phase A} \psfrag{Phase B}[c][c][.7]{Phase B} \psfrag{Phase C}[c][c][.7]{Phase C}
\psfrag{t0}[c][c][.7]{$t_0$} \psfrag{t1}[c][c][.7]{$t_1$} \psfrag{t2}[c][c][.7]{$t_2$}  \psfrag{t3}[c][c][.7]{$t_3$}  \psfrag{t4}[c][c][.6]{$t_4$}

\section{Conclusion}

We have presented a novel guidance and control strategy for capturing and stabilizing a fast-spinning satellite using a spinning-base servicing satellite equipped with a robotic manipulator, joint locks, and reaction wheels. Our approach assumes that the target satellite is spin-stabilized, and that the CoM of the servicing satellite is initially aligned with the axis of rotation of the target. The operation is divided into three sequential phases, each with a specific control objective. In Phase A, we have developed a Lyapunov-based controller for the reaction wheels, which replicates the rotational motion of the target satellite and eliminates the relative motion of the target satellite with respect to the servicing satellite's coordinate frame at its CoM. The matching spin between the two satellites simplifies robot guidance and control in subsequent phases, as only a single trajectory planning is required for the entire operation. In Phase B, we have designed an integrated planning and coordination control scheme to guide the robotic manipulator in capturing the target while maintaining zero relative angular velocity between the two satellites. Finally, in Phase C, we have developed a Lyapunov-based controller for the reaction wheels to damp out the rotation motion of the rigidly interconnected satellites as quickly as possible, subject to actuation torque limits and the maximum torque exerted by the manipulator's end-effector. Simulation results have  demonstrates the effectiveness of our proposed approach for capturing and stabilizing a non-cooperative satellite using the matching spin maneuver concept. At the end of the coordinated control and maneuvers, the angular momentum of the spinning target satellite has been effectively transferred to the reaction wheels for eventual momentum dumping by firing thrusters or magnetorquer rods.

\section*{Appendix}

Using equality constraint \eqref{eq:equl_const} in the inequality constraints \eqref{eq:inequl_const1} and \eqref{eq:inequl_const2}, the latter can be explicitly expressed in terms of the variable $\sigma$ as follows:  
\begin{subequations} 
\begin{align} \label{eq:poly1}
a_1 \sigma^2 + b_1 \sigma + c_1 & \leq 0  \\ \label{eq:poly2}
a_2 \sigma^2 + b_2 \sigma + c_2 & \leq 0
\end{align}
\end{subequations}
where the coefficients of the  polynomials can be found from  
\begin{align*}
a_1&=\bm\omega_b^T \bm M_t^T \bm B^{-T} \bm B^{-1} \bm M_t \bm\omega_b/h^2\\
b_1&=-2 \bm c_t^T \bm B^{-T} \bm B^{-1} \bm M_t \bm\omega_b/h\\
c_1&=\| \bm B^{-1} \bm c_t \|^2 - \tau_{e_{\rm max}}^2\\
a_2&=\bm\omega_b^T \bm M_t^T \bm B^{-T} \bm G^T \bm G \bm B^{-1} \bm M_t \bm\omega_b /h^2\\
b_2&=-2(\bm c_t^T \bm B^{-T} \bm G^T \bm G \bm B^{-1} \bm M_t \bm\omega_b+ \bm c_g^T \bm G \bm B^{-1} \bm M_t \bm\omega_b)/h\\
c_2&=\|\bm G \bm B^{-1} \bm c_t \|^2 + \| \bm c_g \|^2 + 2 \bm c_g^T \bm G \bm B^{-1} \bm c_t - \tau_{e_{\rm max}}^2
\end{align*}
and $h=\| \bm M_t \bm\omega_b \|$ is the magnitude of the angular momentum of the overall system. Suppose $\sigma_1^*$ and $\sigma_2^*$ are the largest roots of the second-order  polynomials in \eqref{eq:poly1} and \eqref{eq:poly2}. Then, one can conclude that the optimal solution to \eqref{eq:maxsigma} is
\begin{equation} \label{eq:sigma}
\sigma = \min(\sigma_1^*, \sigma_2^*)
\end{equation}
\bibliographystyle{IEEEtran}

\end{document}